# ZID-Net: Zero-Inference Diffusion Prior Decoupling Network for Single Image Dehazing★

Xinheng Li, Minghao Chen, Mengqing Wu, Yan Liu*, Guanying Huo*

*College of Information Science and Engineering, Hohai University, Changzhou, 213200, Jiangsu, China*

## Abstract

Single image dehazing is often constrained by a trade-off between restoration quality and computational efficiency. While efficient, CNNs struggle to learn robust priors for dense and non-homogeneous haze. Conversely, diffusion models provide strong generative priors but suffer from severe inference latency and sampling instability. To address these limitations, we propose ZID-Net, a novel framework that explicitly decouples diffusion supervision from feed-forward inference. For efficient inference, we design a frequency-spatial decoupled feed-forward backbone. Within this backbone, a Channel-Spatial Laplacian Mask (CSLM) filters haze-amplified noise to extract purified structural details, while Lightweight Global Context Blocks (LGCBs) establish long-range spatial dependencies to capture global haze variations. A Dynamic Feature Arbitration Block (DFAB) then adaptively fuses these semantic and structural features for robust reconstruction. To provide the backbone with degradation-aware priors without increasing inference cost, we introduce a Zero-Inference Prior Propagation Head (ZI-PPH) during training. ZI-PPH leverages a conditional diffusion process to predict residual noise and regularize

★ This work was supported in part by the National Key Research and Development Program of China under Grant 2023YFB3907203

The project code is available at: https://github.com/XoomitLXH/ZID-Net.

* Corresponding authors.

Email addresses: liuyan_s@163.com (Y. Liu), huoguanying@hhu.edu.cn (G. Huo).

the shared representation. By discarding the diffusion branch at test time, ZID-Net integrates the representation-shaping benefits of diffusion supervision into a pure feed-forward architecture. Extensive experiments on synthetic, real-world, and remote-sensing benchmarks demonstrate robust restoration across diverse haze conditions, with particularly strong performance on StateHaze1k, where ZID-Net achieves an average PSNR gain of 3.06 dB while maintaining efficient deterministic inference at 19.35 ms.

*Keywords:* Image Dehazing, Frequency-spatial decoupling, Zero-inference prior, Diffusion prior, Efficient vision model

## 1. Introduction

Images captured in hazy weather exhibit severe contrast attenuation, color distortion, and detail loss, as suspended particles weaken direct transmission and introduce scattered airlight. These degradations heavily restrict downstream perception tasks in applications such as autonomous driving [1,2], remote sensing [3,4], and medical diagnostics [5,6]. Although the atmospheric scattering model provides a principled physical formulation, recovering a clear image from a single observation remains highly ill-posed. Furthermore, explicitly estimating the transmission map or atmospheric light is frequently unreliable under dense, non-homogeneous, or real-world haze.

Although recent CNN and transformer-based models have advanced single image dehazing, they still struggle with dense and non-homogeneous haze. Specifically, these architectures inherently entangle the underlying scene semantics with haze patterns, making restoration in heavy-haze scenarios highly unstable. Furthermore, when conventional high-frequency priors (e.g., Laplacian filters) are utilized to preserve structural edges, they lack the mechanism to explicitly suppress the noise amplified by the haze. As a result, corrupted details are inadvertently injected back into the restoration process. While transformer and hybrid designs attempt to address this through global context modeling, they significantly increase computational overhead without fundamentally resolving the underlying feature purity problem.

Recently, diffusion models have demonstrated remarkable capability as generative priors for image restoration. Compared to explicit physical formulations such as the atmospheric scattering model [7], diffusion models excel at learning complex, spatially varying haze distributions in real-world scenarios. However, the iterative reverse sampling process incurs massive inference latency and memory overhead, while its stochasticity leads to inference inconsistency and structural artifacts. This creates a fundamental dilemma: feed-forward networks are computationally efficient but struggle to model severe haze, whereas diffusion-based methods are highly expressive yet prohibitively expensive and unstable at test time.

To address this dilemma, we propose ZID-Net, a frequency-spatial decoupled dehazing network guided by a zero-inference diffusion prior. In the feed-forward backbone, Lightweight Global Context Blocks (LGCBs) are embedded at the bottleneck to establish long-range spatial dependencies. Simultaneously, a Channel-Spatial Laplacian Mask (CSLM) dynamically filters haze-amplified noise from color Laplacian residuals, extracting purified chromatic and structural details. A Dynamic Feature Arbitration Block (DFAB) then adaptively fuses the semantic, structural, and upsampled decoding streams for reconstruction. To incorporate generative priors without increasing inference cost, we attach a Zero-Inference Prior Propagation Head (ZI-PPH). During training, this conditional diffusion branch predicts haze-residual noise, explicitly guiding the shared bottleneck to learn degradation-aware features. At test time, the ZI-PPH is entirely discarded, preserving a strictly one-pass inference architecture.

The main contributions of ZID-Net can be summarized as follows:

(1) We propose a framework that decouples diffusion supervision from inference for single image dehazing. By restricting diffusion priors to the training phase, our method improves restoration performance without compromising feed-forward inference speed.

(2) We design a frequency-spatial decoupled architecture for direct inference. This structure integrates LGCBs for global semantic aggregation, CSLM for color Laplacian residual purification, and dynamic feature arbitration to achieve high-quality and efficient image restoration.

(3) We introduce a zero-inference conditional diffusion prior exclusively during the training phase to inject generative supervision into deterministic inference. By predicting haze-residual noise during training, it explicitly guides the shared backbone to model the haze distribution.

## 2. Related Work

Image dehazing has evolved significantly, transitioning from early physical priors to end-to-end deep networks, and most recently, to diffusion models. In this section, we briefly review these three types of methods.

### 2.1. Physical Prior-based Image Dehazing

Early approaches relied heavily on the atmospheric scattering model, seeking to recover the transmission map and global atmospheric light through statistical priors. Representative methods include the Dark Channel Prior (DCP) [8] and the Color Attenuation Prior [9]. However, constrained by their handcrafted nature and simplified physical assumptions, these formulations struggle to adapt to complex real-world hazy conditions. Consequently, they frequently produce severe color distortion and halo artifacts when processing regions with dense haze, bright skies, or substantial depth variations.

### 2.2. End-to-End Image Dehazing

To overcome the fragility of empirical priors, early CNN architectures like DehazeNet [10] and AOD-Net [11] introduced trainable parameter mappings. Subsequent works bypassed physical formulations entirely, advancing to fully end-to-end networks like FFA-Net [12]. Recent models further extend this pipeline with specific auxiliary mechanisms: C2PNet [13] employs contrastive learning to explicitly separate clear and hazy feature representations, DCMPNet [14] reintroduces depth maps to guide physical scene understanding, DEA-Net [15] focuses on restoring high-frequency local details, and SGDN [16] utilizes the YCbCr [17] color space to preserve structural integrity independent of color distortion. Recent methods also explore feature decoupling and knowledge transfer: DCE-Net [18] leverages dual-frequency knowledge for detail and content restoration, ZRID-Net [19] employs self-decoupling and reverse knowledge transfer for zero-reference dehazing, and TSID-Net [20] combines style transfer with contrastive knowledge transfer. While improving visual quality, many learning-based dehazing methods still rely heavily on localized convolutional operations, limiting their ability to model non-homogeneous haze and causing feature entanglement where skip-connections bypass hazy artifacts into the decoder.

To overcome the limited receptive fields of CNNs, Transformers have been widely adopted for global context modeling. For instance, DehazeFormer [21] modifies standard normalization within self-attention to better capture non-local haze distributions. However, since traditional self-attention scales quadratically with image resolution, subsequent models like the MB-TaylorFormer [22,23] series approximate the softmax operation via Taylor expansion to achieve linear complexity. Furthermore, UDPNet [24] employs cross-attention to integrate physical depth cues with visual representations, using depth priors to restore structures under spatially varying haze. Despite these optimizations, Transformer architectures inherently demand higher inference overhead than standard CNNs. To balance computational efficiency with global context, we propose LGCBs. Instead of relying on computationally heavy self-attention variations, LGCBs efficiently establish long-range spatial dependencies, achieving robust global reasoning with the deterministic efficiency of CNNs.

### 2.3. Diffusion Models in Image Dehazing

Diffusion models have recently emerged as a powerful tool in low-level vision by providing strong generative priors. For instance, DiffIR [25] demonstrates the efficacy of diffusion processes in general image restoration. Building on this, recent studies adapt diffusion models specifically for the dehazing task. FGPS [26], for example, integrates frequency-domain guidance into reverse sampling to enhance structural recovery. Recent studies further combine diffusion restoration with optimal-transport-based unpaired training [27] or degradation-aware prompts derived from pretrained vision-language models [28]. Concurrently, generative models are also being exploited to address the discrepancy between synthetic and real-world haze, as seen in methods like *Learning Hazing to Dehazing* [29], which synthesize realistic hazy images for better supervision.

While these approaches advance the field, they face distinct limitations. Employing diffusion models directly for restoration incurs high iterative sampling costs and training-inference inconsistency. Unlike these methods, ZID-Net explores the conditional diffusion branch exclusively as a training regularizer. By predicting residual noise during training, it explicitly guides the shared backbone to model the underlying haze distribution. Because this branch is completely discarded during testing, our method benefits from the diffusion prior, achieving deterministic inference without the latency of iterative sampling.

## 3. Proposed Method

In this section, we detail the architecture of the proposed ZID-Net. As shown in Fig. 1, our framework adopts a decoupled design: a deterministic network for direct inference and an auxiliary diffusion prior branch solely for training. First, Section 3.1 introduces the frequency-spatial decoupled dehazing backbone, which extracts global semantic contexts while preserving high-frequency chromatic details. Next, Section 3.2 presents the Zero-Inference Prior Propagation Head (ZI-PPH). This training-only auxiliary branch utilizes a conditional diffusion process to guide the backbone in modeling the haze distribution, enabling deterministic inference without reverse-sampling latency. Finally, the objective functions are detailed in Section 3.3.

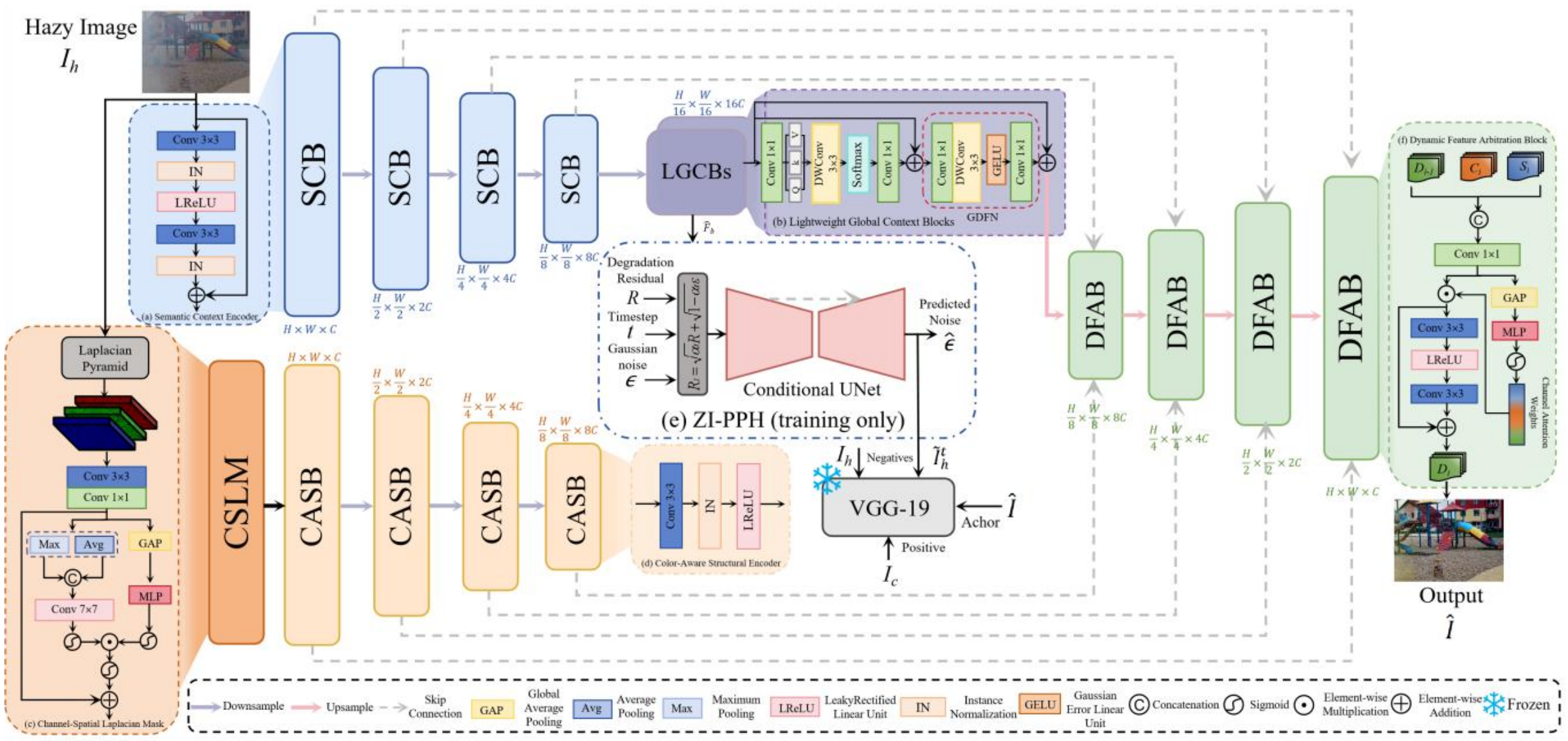

**Fig. 1.** Overall architecture of the proposed ZID-Net. The network adopts a dual-stream deterministic backbone regularized by a diffusion-based prior. (a) The Semantic Context Block (SCB) extracts multi-scale context features from the input hazy image. (b) The LGCBs aggregate scene-level deep semantic features at the bottleneck stage. (c) The CSLM extracts and adaptively filters high-frequency physical residuals using a Laplacian pyramid. (d) The Color-Aware Structural Block (CASB) extracts high-frequency structural features. (e) The ZI-PPH regularizes the bottleneck features via a conditional U-Net during training and is detached during testing. (f) The DFAB fuses multiscale semantic and structural features to decode and reconstruct the final clear image.

### 3.1. Frequency-Spatial Decoupled Dehazing Backbone

As depicted in the main pathway of Fig. 1, the feed-forward backbone employs a dual-stream architecture to process the input. Given a hazy image $I_h$, we first compute a color high-frequency residual $I_{hf}^h$ to decouple the frequency and spatial representations. Then, $I_h$ is fed into a semantic context encoder for global semantic modeling, while $I_{hf}^h$ is routed to a color-aware structural encoder for screened high-frequency extraction. Finally, the DFAB decoder adaptively fuses these dual-branch features with the upsampled states to reconstruct the clear image.

The motivation for this frequency-spatial separation arises from the different spatial behaviors of haze-related variations and scene structures, as illustrated in Fig. 2. Illumination changes and atmospheric effects usually extend over broad regions, resulting in low-frequency variations in image intensity and color. In contrast, scene boundaries, textures, and chromatic transitions occur at local spatial scales and are mainly represented by high-frequency components. Spatially varying haze may also introduce high-frequency fluctuations near regions with abrupt degradation changes, making it necessary to preserve useful structural responses while suppressing unreliable ones. Fig. 2 shows that the high-frequency components of the independent R, G, and B channels maintain strong linear correlations with their clear-image counterparts, whereas a single-channel representation loses part of this chromatic information. These observations motivate using the color high-frequency residual as an independent structural input, while the semantic branch focuses on global haze-related variations.

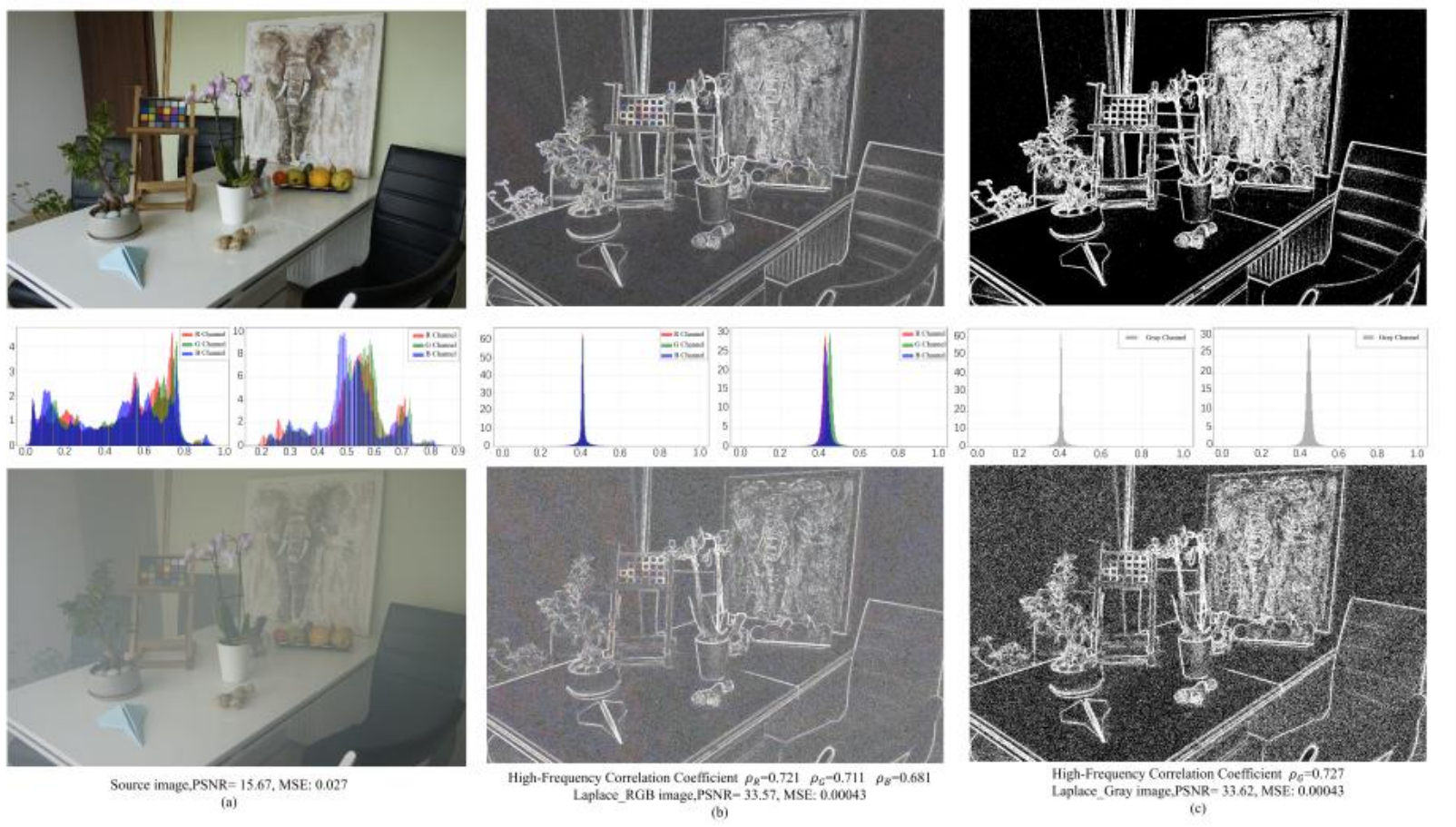

**Fig. 2.** Comparative analysis of high-frequency characteristics between clean and hazy images. (a) Source images with intensity histograms (left: clean; right: hazy) and metrics. (b) RGB-channel Laplacian maps, high-frequency correlations, histograms (left: clean; right: hazy), and metrics. (c) Single-channel Laplacian maps, high-frequency correlations, histograms (left: clean; right: hazy),

and metrics.

**Semantic Context Encoder with LGCB.** Standard convolutions expand the receptive field only gradually, which is restrictive when haze thickness, airlight, and color bias correlate across distant regions. The semantic context encoder therefore applies a $3\times3$ projection with Instance Normalization and LeakyReLU, followed by four stages of a residual block and a stride-2 convolution to produce the multiscale hierarchy $\{S_0,S_1,S_2,S_3\}$. At the deepest stage, the spatial dimensions are compressed to $1/16$ of the original resolution, resulting in a bottleneck feature $F_b \in R^{16C\times\frac{H}{16}\times\frac{W}{16}}$. Defining $N=\frac{HW}{256}$ as the spatial cardinality at this resolution, we use Lightweight Global Context Blocks (LGCBs) on $F_b$ to aggregate scene-level context. This formulation avoids the quadratic computational burden inherent in conventional spatial self-attention.

For an input feature $X\in R^{16C\times\frac{H}{16}\times\frac{W}{16}}$, LGCB first generates the query, key, and value tensors through a fused projection block. Specifically, a point-wise convolution $W_p(\cdot)$ expands the channel dimension to $48C$, followed by a depth-wise convolution $W_d(\cdot)$ to encode spatial context. The resulting tensor is then split equally along the channel dimension:

$$[Q,K,V]=Split(W_d(W_p(X))) \tag{1}$$

where $Q,K,V \in \mathbb{R}^{16C\times\frac{H}{16}\times\frac{W}{16}}$. Rather than computing spatial attention over $N\times N$ locations, LGCB reshapes these tensors into $\mathbb{R}^{16C\times N}$ and performs transposed attention in the channel space. Let $\bar{Q}$ and $\bar{K}$ denote the $l_2$-normalized representations of $Q$ and $K$ along the spatial dimension, and $\bar{V}$ be the reshaped value tensor. The attended representation is computed as:

$$X' = X + W_o\left(Softmax(\bar{K}\bar{Q}^{\top}/\tau)\bar{V}\right) \tag{2}$$

where $\tau$ is a learnable temperature, $\text{Softmax}(\cdot)$ is applied along the channel dimension to compute channel-wise attention weights, and $W_o(\cdot)$ is a $1 \times 1$ output projection.

By computing attention across the $16C$ channel dimensions, the network enables each channel to interact through scene-wide responses rather than being restricted to local windows.

To further refine this mixed representation, LGCB employs a Gated Depth-wise Feed-Forward Network (GDFN). The GDFN expands the channel dimension, applies a depth-wise convolution, and splits the intermediate feature into two parallel paths to perform a gating operation:

$$\widehat{X} = X' + W_2(GELU(U_1) \odot U_2) \quad (3)$$

where $[U_1, U_2] = Split(W_d(W_1(X')))$, and $W_1, W_2$ are point-wise linear projections.

By computing attention across the channel dimension, the computational complexity of LGCB scales linearly with respect to the spatial resolution $N$. Let $C_b$ denote the channel capacity at the bottleneck. The dominant cost of LGCB is $\mathcal{O}(C_b^2 N)$, which provides a significant efficiency advantage over the $\mathcal{O}(N^2 C_b)$ quadratic complexity of conventional spatial self-attention. This linear scaling is particularly advantageous for high-resolution inputs. For example, when increasing the input resolution from $512 \times 512$ to $1024 \times 1024$, the computational overhead of traditional spatial attention scales quadratically, often leading to memory exhaustion. In contrast, LGCB maintains a manageable linear increase. Furthermore, by applying LGCBs exclusively at the maximally compressed bottleneck layer, ZID-Net achieves high-density scene-wide semantic aggregation with minimal computational overhead. The final semantic bottleneck output is denoted by $\widehat{F}_b$.

**Color-Aware Structural Encoder with CSLM.** Motivated by this analysis and the observations in Fig. 2, the Channel-Spatial Laplacian Mask (CSLM) is designed to extract and purify color high-frequency residuals. The module initially extracts a single-scale color residual $I_{hf}^h$ from the hazy input $I_h$:

$$I_{hf}^h = I_h - Up(Down(G(I_h))) \tag{4}$$

where $G(\cdot)$ is a $5 \times 5$ Gaussian filter. Unlike conventional grayscale edge extraction, this operation preserves the channel-specific chromatic structures identified in our analysis.

To map this physical residual into the feature space, CSLM employs a shallow projection followed by a $1 \times 1$ color adaptation layer. To suppress haze-amplified fluctuations while preserving reliable edges, the module functions as a dynamic gating mechanism. It computes a channel descriptor $M_c$ via global average pooling and an MLP, and a spatial descriptor $M_s$ via a $7 \times 7$ convolution over channel-wise pooling responses. The intermediate feature $X$ is then adaptively filtered and reinforced via a residual connection:

$$\widetilde{X} = X + X \odot \sigma(M_c \odot M_s) \tag{5}$$

where $\sigma(\cdot)$ is the sigmoid function. The filtered feature $\widetilde{X}$ then passes through four stride-2 convolutional stages to form the structural hierarchy $\{C_0, C_1, C_2, C_3\}$.

**Dynamic Feature Arbitration Block (DFAB) Decoder.** The decoder reconstructs the clear image by progressively fusing dual-encoder representations with the upsampled decoder states. Starting from the bottleneck $\widehat{F}_b$, each stage employs a DFAB to dynamically arbitrate between the semantic skip feature $S_i$, the structural feature $C_i$, and the upsampled state $\widetilde{D}_i = Up(D_{i+1})$.

Rather than relying on standard addition or concatenation, DFAB performs content-adaptive fusion via channel-wise reweighting. The three-stream input is first concatenated and compressed into a shared representation $Z_i = W_p([\widetilde{D}_i, S_i, C_i])$ via a $1 \times 1$ convolution. DFAB then computes Squeeze-and-Excitation (SE) weights to dynamically recalibrate the channel responses based on the fused content. The recalibrated feature is further refined by a residual block to produce the current decoder state:

$$D_i = ResBlock(Z_i \odot \sigma(W_2 \delta(W_1 GAP(Z_i)))), \quad i \in \{0,1,2,3\} \tag{6}$$

where $GAP(\cdot)$ is global average pooling, and $\delta(\cdot)$ denotes the ReLU activation.

By utilizing SE-style channel reweighting and residual refinement independently at each scale, DFAB balances structural and semantic information effectively, avoiding the computational overhead of explicit spatial or cross-attention mechanisms before the final 3×3 prediction layer.

The semantic and structural branches play complementary roles in image reconstruction. The semantic branch models global illumination and long-range context at the bottleneck, whereas the Laplacian branch provides local edges, textures, and color details. CSLM filters unreliable high-frequency responses caused by haze, and DFAB adaptively combines the semantic, structural, and decoder features at each scale to reconstruct regions with different haze conditions.

### 3.2. Zero-Inference Prior Propagation Head

Methods grounded in the atmospheric scattering model [10,11] often suffer from error accumulation when explicitly estimating the transmission map and atmospheric light under restrictive physical assumptions. Conversely, while purely data-driven feed-forward dehazers bypass these physical bottlenecks and offer high computational efficiency, relying entirely on standard reconstruction losses leaves their degradation modeling weakly constrained, particularly under dense or spatially varying haze. While diffusion models offer strong generative priors, their direct application incurs heavy iterative sampling costs. To bypass this overhead, we introduce a training-only auxiliary branch termed the Zero-Inference Prior Propagation Head (ZI-PPH). As illustrated in Fig. 3, this module operates entirely within the haze-residual domain.

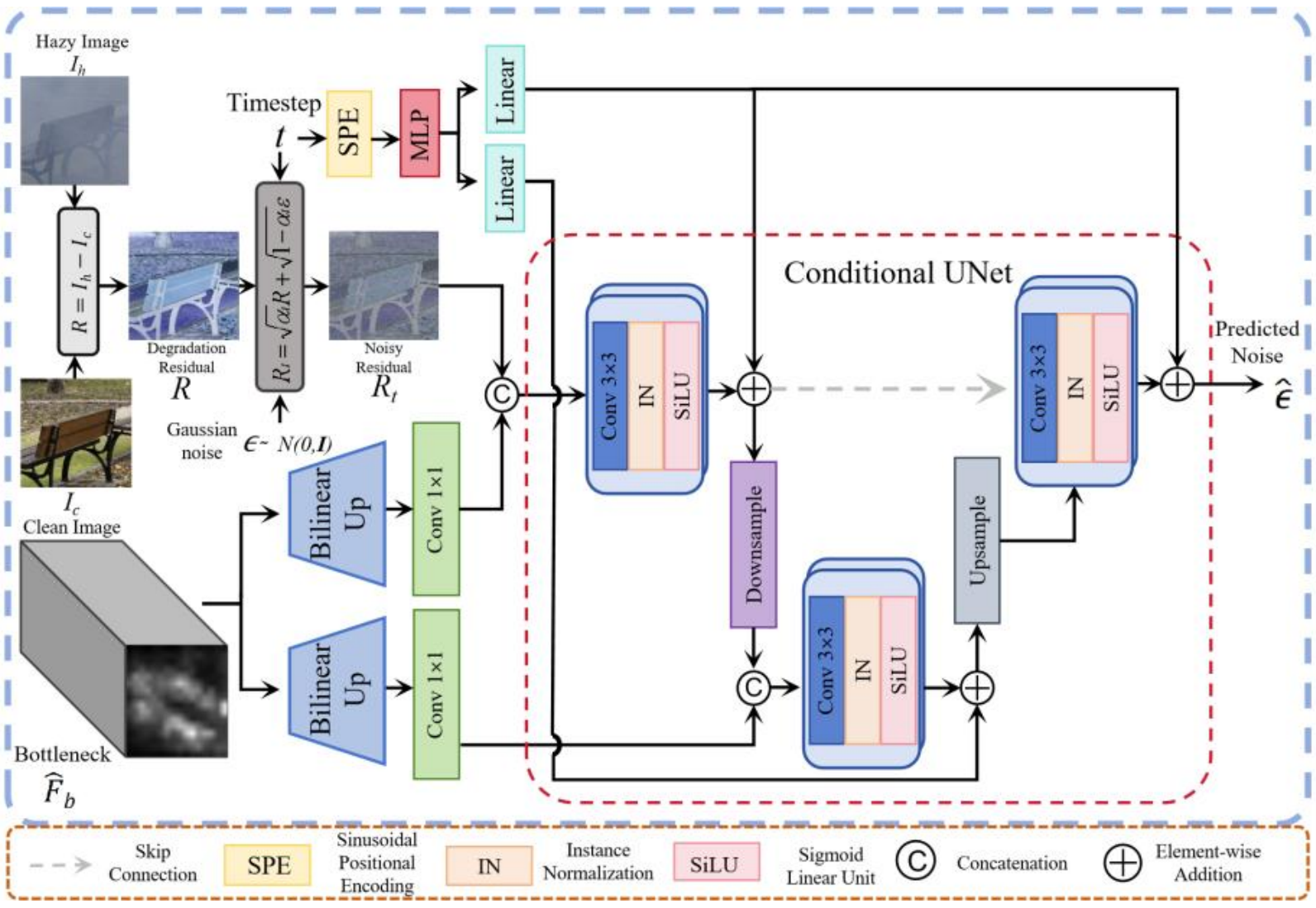

**Fig. 3.** Detailed architecture of the ZI-PPH branch. This training-only auxiliary module regularizes

the shared bottleneck features $\widehat{F}_b$ via a conditional U-Net. The noisy physical residual $R_t$ and $\widehat{F}_b$ are concatenated at the input level, with the timestep $t$ (encoded via SPE and MLP) additively injected across the network. This branch predicts the added noise $\widehat{\epsilon}$ to provide auxiliary degradation-aware supervision to the shared representation.

Given a hazy-clean pair $(I_h, I_c)$, we first define the degradation residual as $R = I_h - I_c$. For a randomly sampled diffusion timestep $t$, the forward perturbation process constructs a noisy residual:

$$R_t = \sqrt{\alpha_t}R + \sqrt{1 - \alpha_t}\epsilon \tag{7}$$

where $\epsilon \sim \mathcal{N}(0, \mathbf{I})$ and $\alpha_t$ is the cumulative product of the variance schedule.

Unlike conventional diffusion models that sample $t$ uniformly from $[0, T]$, we introduce a batch-wise severity-adaptive sampling strategy. For the $i$-th sample in a mini-batch, we calculate the mean absolute error of its residual, $s_i = mean(|I_{h,i} - I_{c,i}|)$, to estimate the haze concentration. This is normalized within the batch to yield a relative severity score $s_{norm,i} \in [0,1]$. Using this score, we dynamically constrain the upper bound of the sampled timestep:

$$T_i^{cap} = T_{low} + \gamma \cdot s_{norm,i} \cdot (T - T_{low}) \tag{8}$$

where $T_{low}$ is a predefined baseline and $\gamma$ controls the adaptive strength. The timestep is then uniformly drawn from $[0, T_i^{cap}]$. Consequently, lighter haze samples are assigned smaller timesteps to preserve high-frequency details, while heavier haze samples are exposed to a broader noise range to capture robust structural priors.

Rather than executing a full reverse chain, ZI-PPH predicts the injected perturbation $\epsilon$ conditioned on the shared semantic bottleneck representation $\widehat{F}_b$. This conditioning turns the diffusion objective into an auxiliary feature-level regularizer: accurate perturbation estimation across varying timesteps requires $\widehat{F}_b$ to preserve structure-consistent semantic information while remaining informative about haze-dependent degradation. To transfer this degradation-aware guidance effectively through the conditional U-Net, $\widehat{F}_b$ is spatially aligned and fused at two distinct semantic scales.

At the input level, $\widehat{F}_b$ is bilinearly upsampled to the original resolution, compressed via a $1 \times 1$ projection, and concatenated with the noisy residual $R_t$. This early injection provides pixel-aligned global context, enabling the shallow convolutional layers to distinguish between the spatial haze distribution and the underlying structural details.

Deeper in the network, $\widehat{F}_b$ is upsampled to the half-resolution scale and additively integrated into the intermediate encoder representations. This secondary fusion reinforces the semantic prior at a higher level of abstraction, preventing the conditioning signal from decaying as features undergo progressive spatial downsampling.

Alongside this multi-scale spatial conditioning, the diffusion timestep $t$ is encoded via sinusoidal embeddings and a multi-layer perceptron, then added across the U-Net hierarchy. This temporal modulation ensures that all layers maintain awareness of the current perturbation level along the denoising trajectory. Formally, the auxiliary branch predicts:

$$\widehat{\epsilon} = \epsilon_\theta(R_t, t, \widehat{F}_b) \tag{9}$$

where $\widehat{\epsilon}$ denotes the predicted noise at timestep $t$, and $\epsilon$ is the ground-truth Gaussian noise added during the forward diffusion process. Here, $\epsilon_\theta$ represents the conditional U-Net of the auxiliary branch. Since the residual $R = I_h - I_c$ explicitly represents the degradation introduced by haze, diffusion is performed directly in the degradation residual space. At each timestep, $R_t$ is a perturbed version of $R$, and predicting the added noise provides the denoising supervision needed to estimate the underlying haze residual from different noise levels. Training across multiple timesteps therefore encourages the conditional U-Net to learn the statistical characteristics of haze residuals, including their degradation strength and spatial variation. Because noise prediction is conditioned on $\widehat{F}_b$, the diffusion loss is backpropagated through ZI-PPH to the S-Encoder and LGCB modules. As a result, haze-dependent feature deviations that hinder consistent residual prediction across different noise levels are progressively penalized during training, encouraging the shared bottleneck to form a more structure-consistent and degradation-aware representation. This implicit regularization reduces the influence of haze-induced feature shifts without requiring an explicit feature-distribution alignment loss, while preserving structural and semantic information that is also beneficial to image reconstruction in the main decoder.

Notably, ZI-PPH introduces no inference-time overhead. During testing, the ZI-PPH branch is detached, and the network reconstructs the clear image via a single forward pass.

$$\widehat{I} = f_{\theta}(I_h, I_{hf}^h) \tag{10}$$

where $\widehat{I}$ denotes the final predicted clear image, and $f_{\theta}$ represents the deterministic feed-forward dehazing backbone parameterized by $\theta$. This formulation bypasses all residual noising steps, conditional U-Net evaluations, and reverse diffusion chains. Thus, ZI-PPH transfers the representation-shaping benefits of diffusion supervision to the shared backbone during training, while maintaining the deterministic efficiency of a feed-forward architecture.

### 3.3. Training Objectives

The overall training objective of ZID-Net is a weighted combination of three terms:

$$\mathcal{L}_{total} = \lambda_1\mathcal{L}_1 + \lambda_2\mathcal{L}_{contrast} + \lambda_3\mathcal{L}_{diff} \tag{11}$$

where the empirically set hyperparameters are $\lambda_1 = 1.0$, $\lambda_2 = 0.1$, and $\lambda_3 = 0.35$. The reconstruction term employs a standard pixel-wise $L_1$ loss to constrain the fidelity between the predicted clear image $\widehat{I}$ and the ground truth $I_c$.

$$\mathcal{L}_1 = \|\widehat{I} - I_c\|_1 \tag{12}$$

To enhance perceptual quality, we utilize a contrastive loss computed in the VGG-19 feature space. It pulls the restored image closer to the clean target (positive) while pushing it away from both the hazy input and a degradation-focused noisy residual (negatives). Let $\phi_l(\cdot)$ denote the extracted features from the $l$-th layer of a pre-trained VGG-19 network [30], with corresponding weights $\omega_l$. Defining the clipped noisy residual image as $\widetilde{I}_h^t = \text{clip}(I_c + R_t, 0,1)$, the contrastive loss is formulated as:

$$\mathcal{L}_{contrast} = \sum_{l} \omega_l \frac{\|\phi_l(\widehat{I}) - \phi_l(I_c)\|_1}{\|\phi_l(\widehat{I}) - \phi_l(I_h)\|_1 + \|\phi_l(\widehat{I}) - \phi_l(\widetilde{I}_h^t)\|_1 + \varepsilon} \tag{13}$$

where $\varepsilon = 10^{-7}$ prevents numerical instability. Here, $I_h$ acts as an image-domain negative, while $\widetilde{I}_h^t$ serves as a degradation-focused negative explicitly tied to the diffusion perturbation.

Finally, the diffusion supervision term penalizes the $L_1$ distance between the predicted noise $\widehat{\epsilon}$ and the sampled Gaussian noise $\epsilon$:

$$\mathcal{L}_{diff} = \|\widehat{\epsilon} - \epsilon\|_1 \tag{14}$$

Together, these three losses tightly couple pixel-level reconstruction, perceptual discrimination, and generative degradation modeling into a single, cohesive optimization framework.

# 4. Experiments

In this section, we comprehensively evaluate the effectiveness and generalization of our ZID-Net. Section 4.1 describes the experimental setup, including datasets, evaluation metrics, and baselines. Then, we assess our model on synthetic haze in Section 4.2, real-world conditions in Section 4.3, and remote sensing imagery in Section 4.4. Section 4.5 analyzes the computational efficiency of our deterministic architecture, and Section 4.6 presents ablation studies to validate the core components.

## 4.1. Experimental Setup

**Implementation details.** All experiments are implemented in PyTorch and conducted on a single NVIDIA GeForce RTX 5060 Ti GPU with 16 GB of VRAM. The network is trained using the Adam optimizer with default momentum parameters and a batch size of 8. The initial learning rate is set to 0.0001 and gradually decays via cosine annealing. To augment the training data, we randomly crop the images into 256×256 spatial patches, followed by random rotations, flips, and mild scaling. For the ZI-PPH branch, the diffusion prior employs $T = 1000$ forward steps. The contrastive loss is computed using multi-scale features extracted from a pre-trained VGG-19 network [30].

**Datasets.** We evaluate our model across synthetic, real-world, and remote sensing scenarios. For synthetic haze, we use the RESIDE [31] benchmark, training on ITS and OTS, and testing on the corresponding SOTS-Indoor and SOTS-Outdoor sets. To evaluate robustness across diverse real-world scenes, we establish the Real-World Thin setting by merging NH-HAZE [32], I-HAZE [33], and O-HAZE [34]. This provides 155 pairs covering indoor and outdoor scenes with homogeneous and non-homogeneous haze, which are partitioned into 145 training and 10 testing pairs. For extreme conditions, the Real-World Dense setting utilizes the Dense-Haze [35] dataset with 50 training and 5 testing pairs to assess structural recovery under severe physical degradation. Finally, for remote sensing, we employ the StateHaze1k [36] dataset, which comprises Thin, Moderate, and Thick subsets, each providing 320 training and 45 testing pairs.

**Evaluation metrics.** We quantitatively assess restoration quality using Peak Signal-to-Noise Ratio (PSNR) [37] and Structural Similarity Index (SSIM) [38]. To evaluate the computational efficiency of our feed-forward architecture, we report the trainable parameters, GFLOPs, and average inference runtime in milliseconds. Both FLOPs and runtime are measured at 256×256 and 512×512 resolutions to demonstrate the linear scaling complexity.

**Baselines.** We compare ZID-Net against twelve state-of-the-art dehazing methods: the physical prior-based DCP [8]; CNN-based models including FFA [12], SGID [39], C2PNet [13], SANet [40], DCMPNet [14], DEA-Net [15], SGDN [16], and TUR [41]; Transformer-based architectures including DehazeFormer [21] and MB-TaylorFormerV2 [23]; and the diffusion-based *Learning Hazing to Dehazing* [29]. For qualitative visual comparisons, we select a subset of these baselines: DCP [8], C2PNet [13], DCMPNet [14], SGDN [16], DehazeFormer [21], MB-TaylorFormerV2 [23], and *Learning Hazing to Dehazing* [29].

### 4.2. Evaluation on Synthetic Datasets

We first assess the restoration capabilities on the synthetic RESIDE benchmark. Quantitatively, as detailed in Table 1, ZID-Net establishes state-of-the-art performance on the synthetic RESIDE benchmark. On the SOTS-Indoor dataset, our model reaches a PSNR of 42.77 dB and an SSIM of 0.997. This advantage continues on the SOTS-Outdoor dataset, where ZID-Net exceeds the second-best method by 0.72 dB, achieving a PSNR of 38.73 dB and an SSIM of 0.993. This margin confirms that our frequency-spatial decoupled backbone effectively preserves structural and chromatic details under synthetic degradation.

**Table 1**

Quantitative comparison results with representative dehazing methods on SOTS-Indoor and SOTS-Outdoor. For each metric, the best results are highlighted in red and second-best in blue.

| Method | Venue & Year | SOTS-Indoor | | SOTS-Outdoor | | Average | |
|---|---|---|---|---|---|---|---|
| | | PSNR↑ | SSIM↑ | PSNR↑ | SSIM↑ | PSNR↑ | SSIM↑ |
| DCP | TPAMI 2010 | 19.50 | 0.859 | 16.98 | 0.858 | 18.24 | 0.859 |
| FFA | AAAI 2020 | 36.39 | 0.989 | 33.57 | 0.984 | 34.98 | 0.986 |
| SGID | TIP 2022 | 38.52 | 0.991 | 34.36 | 0.983 | 36.44 | 0.987 |
| C2PNet | CVPR 2023 | 42.56 | 0.995 | 36.68 | 0.990 | 39.62 | 0.993 |
| Dehaze Former-B | TIP 2023 | 37.84 | 0.994 | 34.95 | 0.984 | 36.40 | 0.989 |
| SANet | IJCAI 2023 | 40.40 | 0.996 | 38.01 | 0.995 | 39.21 | 0.996 |
| DCMPNet | CVPR 2024 | 42.18 | 0.997 | 36.56 | 0.993 | 39.37 | 0.995 |
| DEA-Net | TIP 2024 | 41.31 | 0.995 | 36.59 | 0.990 | 38.95 | 0.993 |
| MB-TaylorFormerV2-B | TPAMI 2025 | 41.00 | 0.993 | 37.81 | 0.991 | 39.41 | 0.992 |
| TUR | AAAI 2025 | 31.17 | 0.976 | 28.75 | 0.972 | 29.96 | 0.974 |
| SGDN | AAAI 2025 | 35.56 | 0.985 | 31.92 | 0.981 | 33.74 | 0.983 |
| *Learning Hazing to Dehazing* | CVPR 2025 | 14.77 | 0.540 | 17.31 | 0.588 | 16.04 | 0.564 |
| **Ours** | - | 42.77 | 0.997 | 38.73 | 0.993 | 40.75 | 0.995 |

Consistent with the quantitative metrics, the qualitative evaluations validate our architectural advantages. As visualized in Figs. 4 and 5, the traditional DCP frequently exhibits color shifts and illumination imbalances. In contrast, deterministic deep learning methods demonstrate superior restoration consistency, effectively removing synthetic haze. However, a closer inspection of the magnified regions reveals that the diffusion-based *Learning Hazing to Dehazing* [29] approach exhibits artifacts stemming from the inherent stochasticity of its sampling process. ZID-Net avoids these generative instabilities. By employing a deterministic inference architecture, our method provides stable dehazing performance comparable to advanced baselines while maintaining higher color fidelity and sharper structural boundaries.

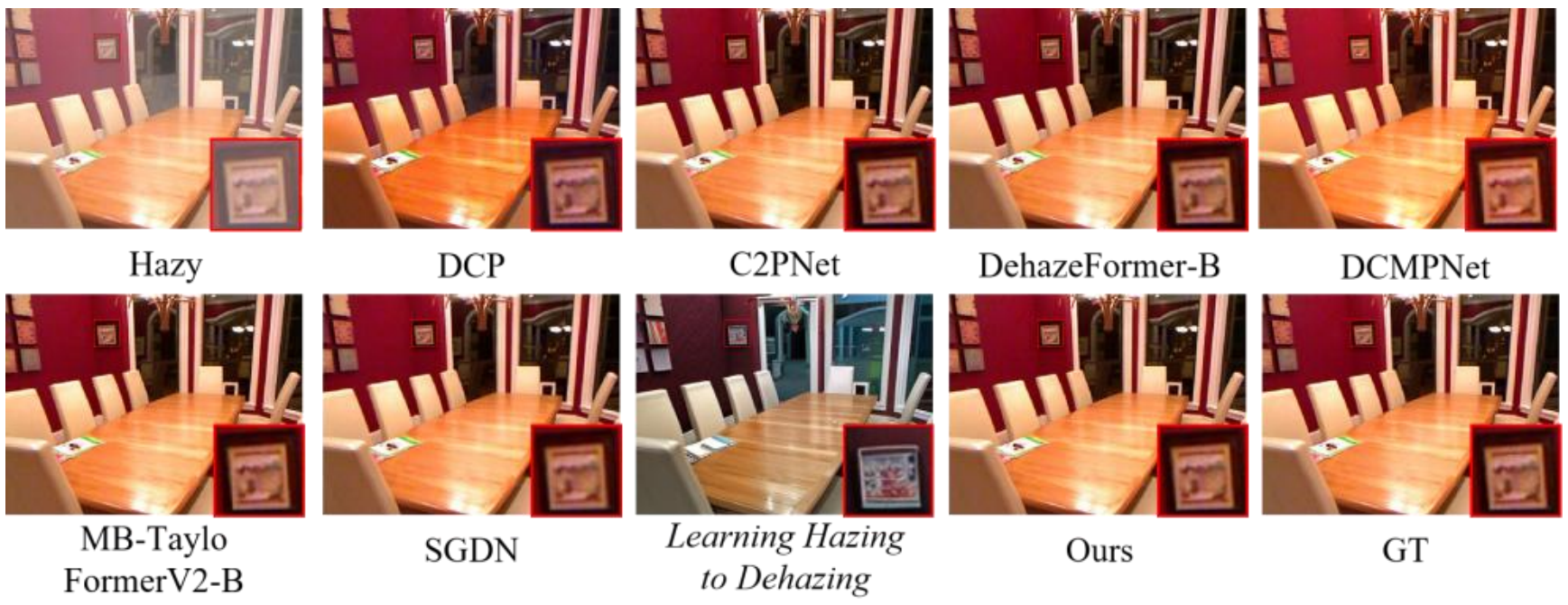

**Fig. 4.** Comparison results of the proposed method with each method on SOTS-Indoor dataset.

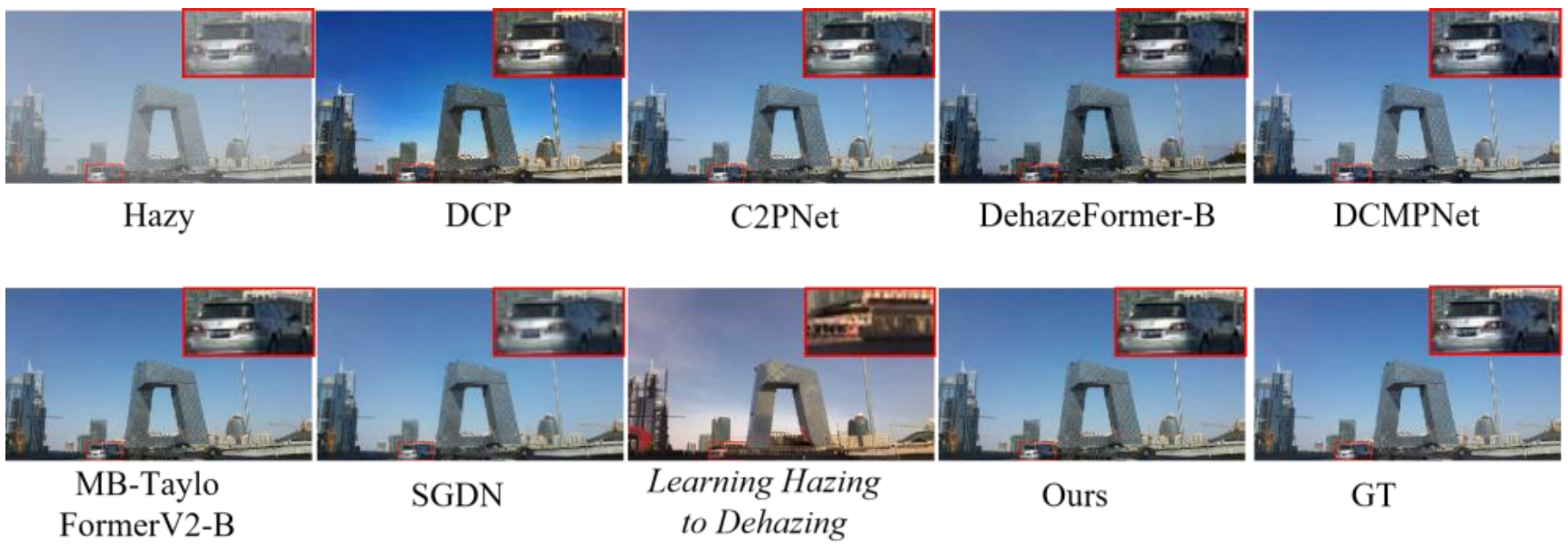

**Fig. 5.** Comparison results of the proposed method with each method on SOTS-Outdoor dataset.

## 4.3. Evaluation on Real-World Datasets

To verify generalization in practical scenarios, Table 2 reports the quantitative results on the Real-World Thin and Dense datasets. Standard diffusion models generally underperform on pixel-aligned metrics like PSNR and SSIM because their generative formulation favors perceptual quality and distribution alignment over deterministic spatial fidelity. ZID-Net circumvents this limitation by utilizing a deterministic inference architecture guided by diffusion priors. On the Real-World Thin dataset, our method outperforms the second-best approach by 1.37 dB in PSNR and 0.044 in SSIM. For the Real-World Dense dataset, which contains scenes with severe haze and substantial visibility obstruction, ZID-Net achieves a 0.97 dB PSNR improvement over the previous state-of-the-art, together with an SSIM of 0.601. This result demonstrates its ability to maintain effective structural perception and pixel-level restoration even under heavily obscured conditions.

**Table 2**

Quantitative comparison results with excellent methods on Real-World Thin and Real-World Dense. For each metric, the best results are highlighted in red and second-best in blue.

| Method | Venue & Year | Real-World Thin | | Real-World Dense | | Average | |
|---|---|---|---|---|---|---|---|
| | | PSNR↑ | SSIM↑ | PSNR↑ | SSIM↑ | PSNR↑ | SSIM↑ |
| DCP | TPAMI 2010 | 14.46 | 0.599 | 13.43 | 0.465 | 13.95 | 0.532 |
| FFA | AAAI 2020 | 15.86 | 0.637 | 18.80 | 0.577 | 17.33 | 0.607 |
| SGID | TIP 2022 | 17.47 | 0.681 | 15.57 | 0.558 | 16.52 | 0.620 |
| C2PNet | CVPR 2023 | 18.83 | 0.708 | 16.86 | 0.558 | 17.85 | 0.633 |
| Dehaze Former-B | TIP 2023 | 16.65 | 0.692 | 18.49 | 0.602 | 17.57 | 0.647 |
| SANet | IJCAI 2023 | 17.42 | 0.724 | 18.11 | 0.595 | 17.77 | 0.660 |
| DCMPNet | CVPR 2024 | 15.31 | 0.653 | 13.32 | 0.149 | 14.32 | 0.401 |
| DEA-Net | TIP 2024 | 18.53 | 0.719 | 19.52 | 0.533 | 19.03 | 0.626 |
| MB-TaylorFormerV2-B | TPAMI 2025 | 19.37 | 0.702 | 19.40 | 0.593 | 19.39 | 0.648 |
| TUR | AAAI 2025 | 19.46 | 0.721 | 12.65 | 0.499 | 16.06 | 0.610 |
| SGDN | AAAI 2025 | 20.46 | 0.732 | 19.90 | 0.623 | 20.18 | 0.678 |
| *Learning Hazing to Dehazing* | CVPR 2025 | 15.02 | 0.393 | 13.21 | 0.431 | 14.12 | 0.412 |
| **Ours** | - | 21.83 | 0.776 | 20.87 | 0.601 | 21.35 | 0.689 |

The visual advantages of ZID-Net are evident under authentic degradation conditions. A closer inspection of the magnified floral regions in Fig. 6 reveals that convolutional networks such as C2PNet [13], DCMPNet [14], and SGDN [16] exhibit overexposure, which leads to the loss of fine-grained petal textures and structural details. In contrast, ZID-Net maintains balanced illumination and preserves these high-frequency components. Furthermore, as shown in Fig. 7, competing methods exhibit chromatic distortion and fail to recover intrinsic colors under dense haze. In contrast, the scene-level global perception provided by LGCBs enables ZID-Net to maintain balanced local exposure, preventing color shifts during real-world image restoration.

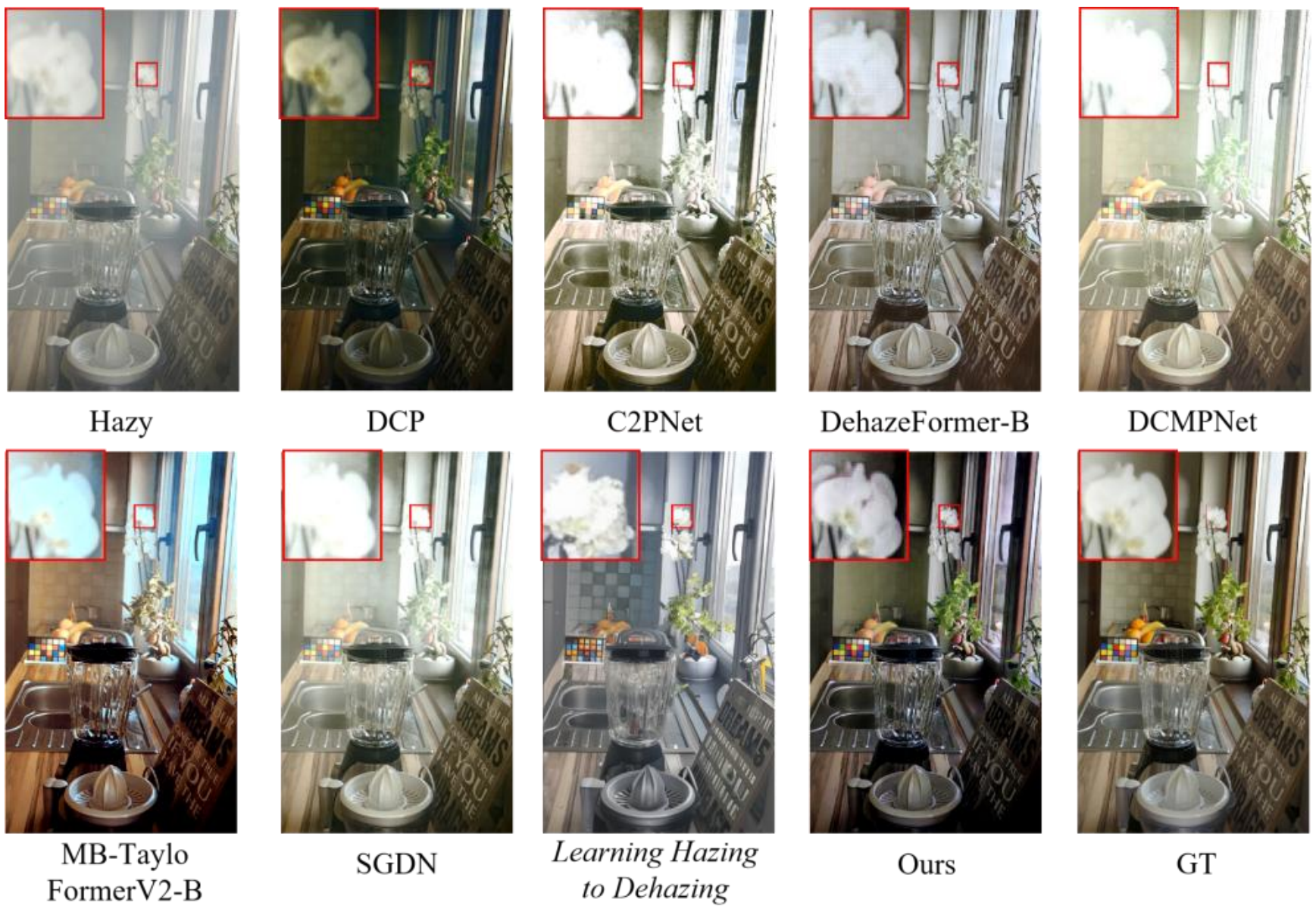

**Fig. 6.** Comparison results of the proposed method with each method on Real-World Thin dataset.

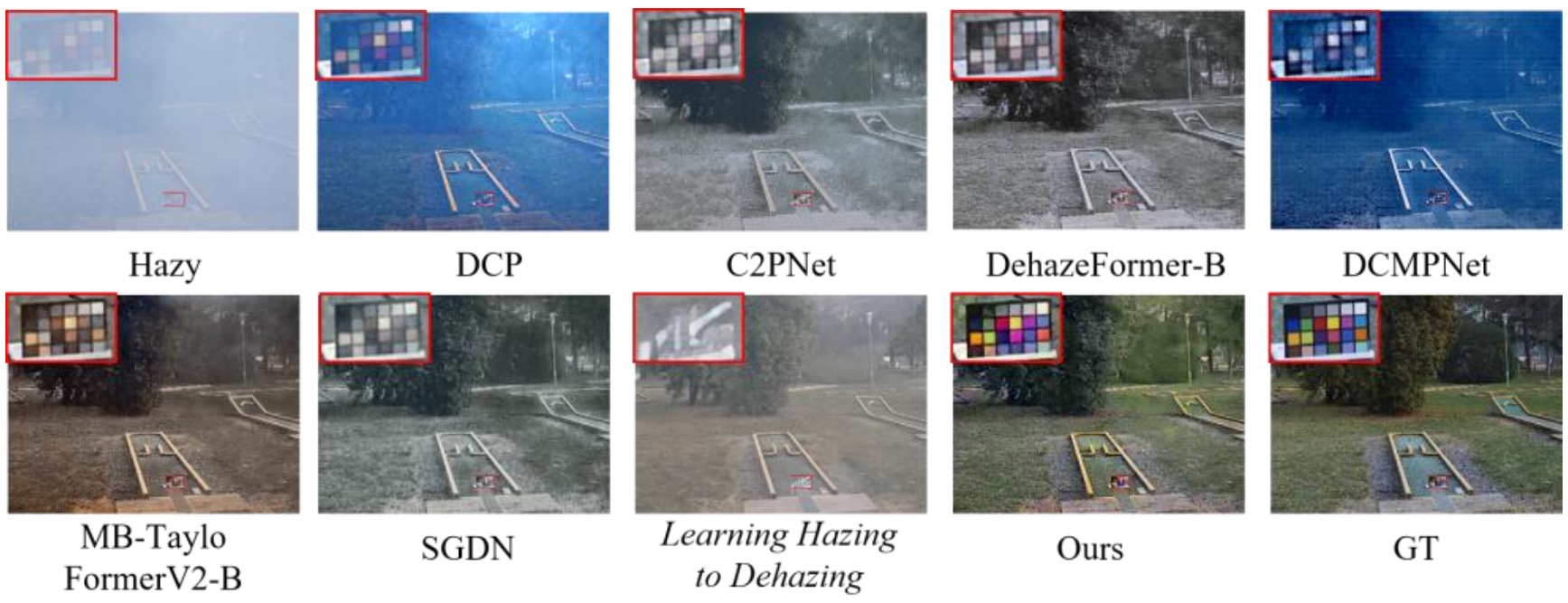

**Fig. 7.** Comparison results of the proposed method with each method on Real-World Dense dataset.

## 4.4. Evaluation on Remote Sensing Imagery

We further extend the evaluation to the remote sensing domain using the StateHaze1k dataset. As reported in Table 3, our prior-guided deterministic architecture establishes a consistent advantage on the StateHaze1k dataset across varying density levels. By utilizing the diffusion prior to inject severity-aware structural guidance into the deterministic backbone, ZID-Net achieves PSNR improvements of 1.16 dB and 1.36 dB over the second-best approaches under Thin and Moderate conditions. The architectural benefit becomes particularly evident under Thick haze, where severe spatial degradation typically limits conventional feature extraction. In this highly degraded scenario, the diffusion prior regularizes the feature space, allowing our method to surpass the next best performance by 5.57 dB in PSNR and 0.061 in SSIM. Overall, ZID-Net achieves the highest average PSNR of 25.29 dB and an SSIM of 0.915, leading the closest baseline by an average of 3.06 dB in PSNR and 0.032 in SSIM. These quantitative gains confirm that integrating generative priors into a feed-forward network robustly mitigates complex remote sensing degradation.

**Table 3**

Quantitative evaluation results on the remote sensing dataset StateHaze1k under different haze densities (Thin, Moderate, and Thick). For each metric, the best results are highlighted in red and second-best in blue.

| Method | Venue & Year | Thin | | Moderate | | Thick | | Average | |
|---|---|---|---|---|---|---|---|---|---|
| | | PSNR↑ | SSIM↑ | PSNR↑ | SSIM↑ | PSNR↑ | SSIM↑ | PSNR↑ | SSIM↑ |
| DCP | TPAMI 2010 | 20.61 | 0.886 | 21.55 | 0.919 | 16.73 | 0.773 | 19.63 | 0.859 |
| FFA | AAAI 2020 | 21.29 | 0.895 | 24.23 | 0.926 | 12.14 | 0.648 | 19.22 | 0.823 |
| SGID | TIP 2022 | 21.69 | 0.906 | 23.68 | 0.940 | 12.29 | 0.692 | 19.22 | 0.846 |
| C2PNet | CVPR 2023 | 22.66 | 0.909 | 25.76 | 0.940 | 14.66 | 0.761 | 21.02 | 0.870 |
| Dehaze Former-B | TIP 2023 | 24.10 | 0.916 | 26.26 | 0.940 | 15.83 | 0.751 | 22.06 | 0.869 |
| SANet | IJCAI 2023 | 22.69 | 0.907 | 24.86 | 0.940 | 17.34 | 0.766 | 21.63 | 0.871 |
| DCMPNet | CVPR 2024 | 22.87 | 0.905 | 24.09 | 0.912 | 14.83 | 0.728 | 20.60 | 0.849 |
| DEA-Net | TIP 2024 | 23.73 | 0.903 | 25.72 | 0.923 | 17.24 | 0.789 | 22.23 | 0.872 |
| MB-TaylorF ormerV2-B | TPAMI 2025 | 21.12 | 0.898 | 24.62 | 0.944 | 16.52 | 0.808 | 20.75 | 0.883 |
| TUR | AAAI 2025 | 23.28 | 0.913 | 26.35 | 0.939 | 14.55 | 0.771 | 21.39 | 0.874 |
| SGDN | AAAI 2025 | 22.67 | 0.907 | 25.69 | 0.940 | 13.32 | 0.735 | 20.56 | 0.861 |
| *Learning Hazing to Dehazing* | CVPR 2025 | 13.79 | 0.217 | 15.55 | 0.214 | 13.49 | 0.196 | 14.28 | 0.209 |
| **Ours** | - | 25.26 | 0.928 | 27.71 | 0.947 | 22.91 | 0.869 | 25.29 | 0.915 |

The visual comparisons directly corroborate these quantitative gains. Figs. 8, 9 and 10 demonstrate the results across thin, moderate, and thick haze conditions, respectively. As the haze density increases, baseline methods such as DehazeFormer [21] and MB-TaylorFormerV2 [23] degrade significantly. They fail to fully remove the haze layer, leaving residual artifacts that obscure geographical features. The magnified regions in Fig. 10 further reveal that DehazeFormer [21] introduces severe color distortion under thick, non-homogeneous haze. In contrast, ZID-Net preserves sharp edges and structural details across different haze densities, including regions with severe local obscuration, and reliably reconstructs surface textures under highly restricted visibility. This robustness highlights the advantage of the dual-stream architecture. By using an independent high-frequency stream, the network can infer and recover structural information that is weakened or partially hidden by dense, spatially varying haze.

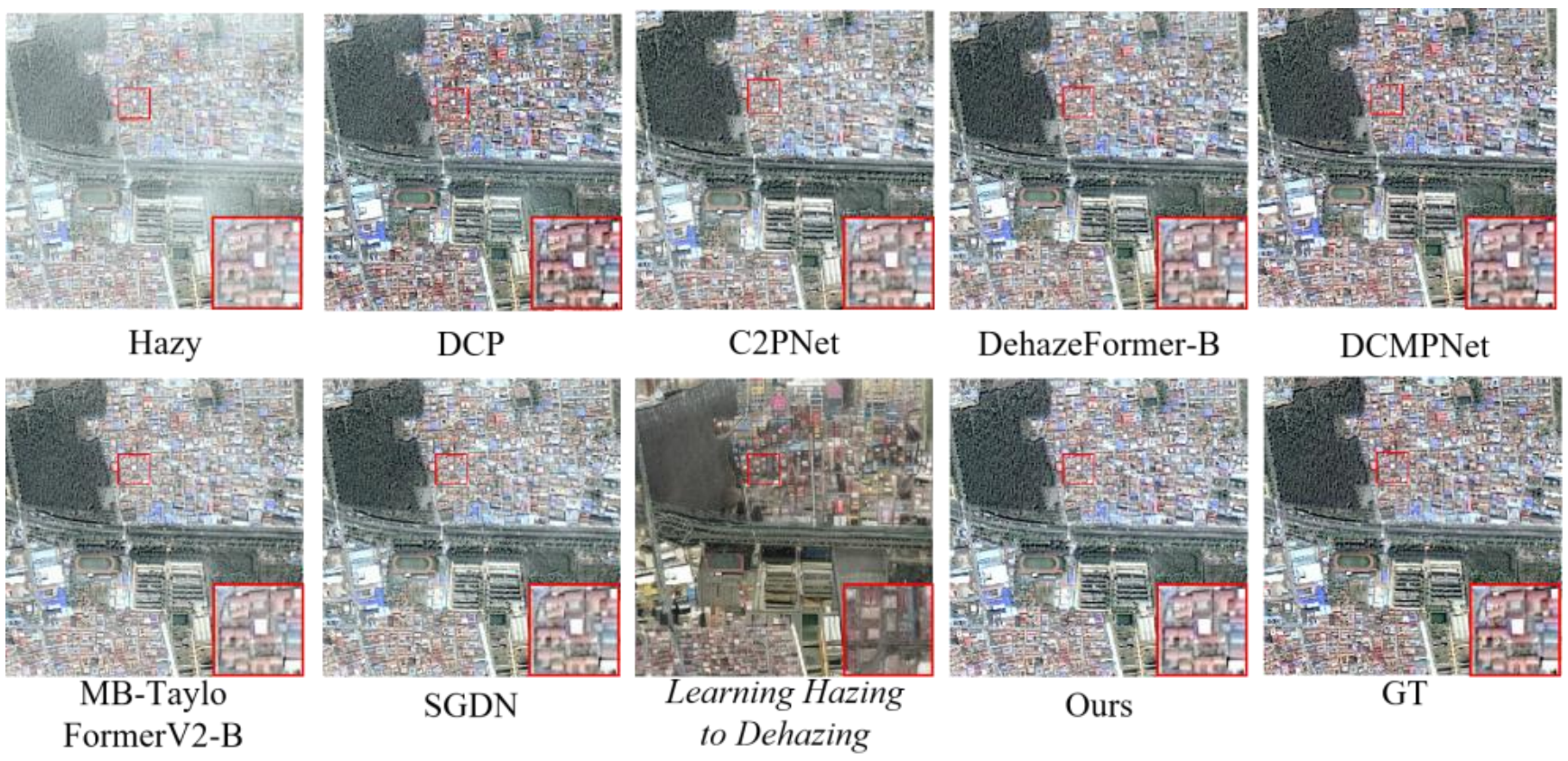

**Fig. 8.** Comparison results of the proposed method with each method on StateHaze1k Thin dataset.

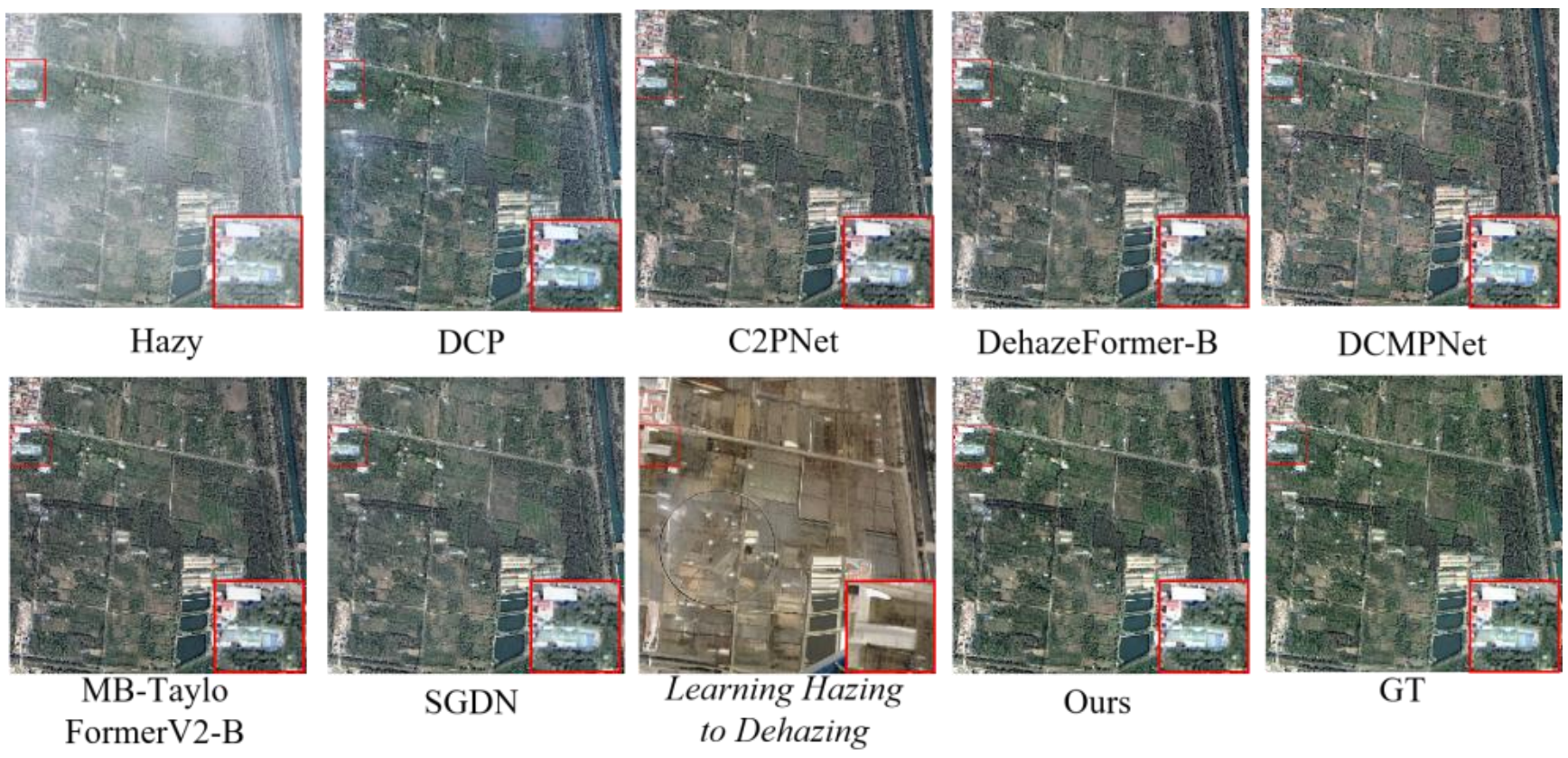

**Fig. 9.** Comparison results of the proposed method with each method on StateHaze1k Moderate dataset.

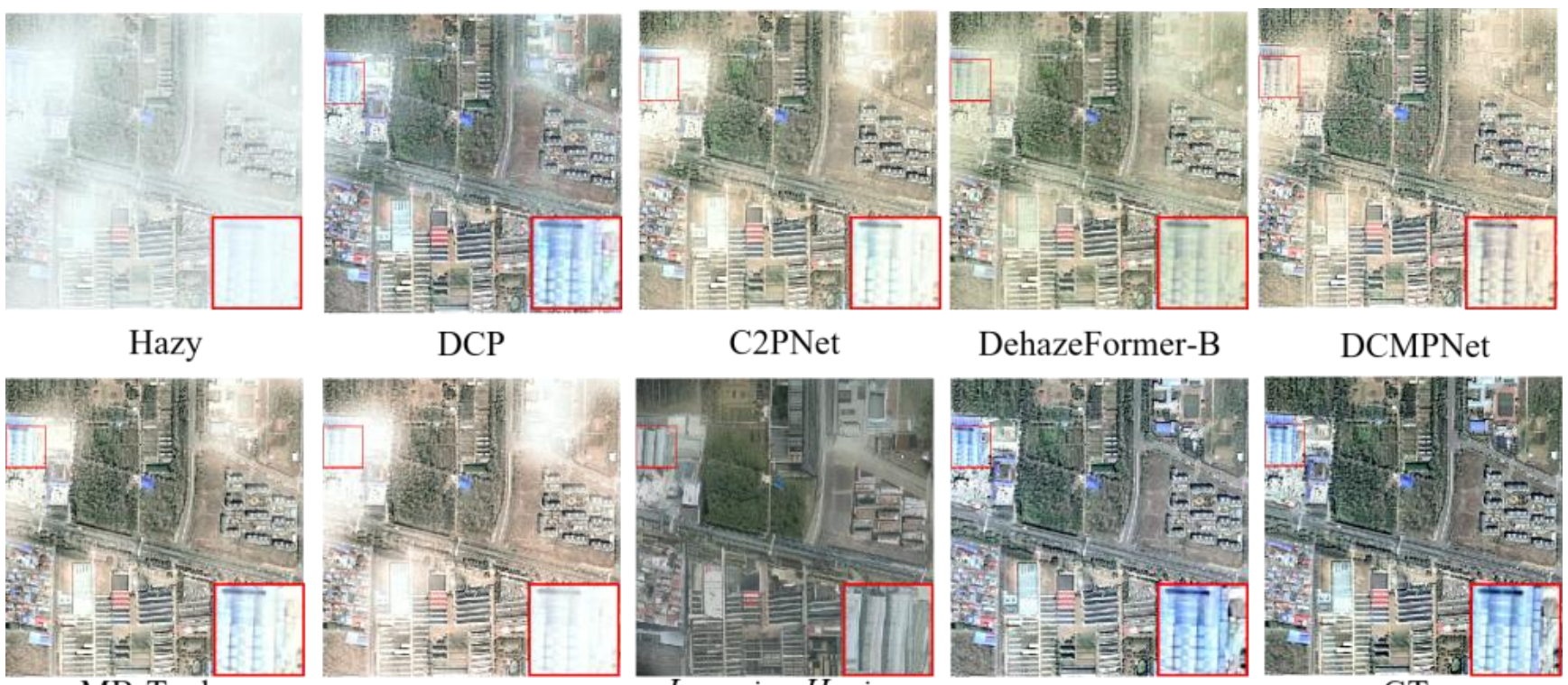

**Fig. 10.** Comparison results of the proposed method with each method on StateHaze1k Thick dataset.

### 4.5. Computational Efficiency Analysis

To evaluate practical applicability, we analyze the computational efficiency across 256×256 and 512×512 spatial resolutions, as detailed in Tables 4 and 5. Notably, the 512×512 evaluation aligns with the native resolution of the StateHaze1k dataset, directly reflecting the processing demands of remote sensing imagery. While ZID-Net maintains a moderate parameter count of 17.51 M to support its dual-stream feature extraction, it remains computationally efficient, requiring only 17.91 GMac and 71.64 GMac at the respective resolutions. The primary architectural advantage, however, emerges in inference latency. Because the diffusion prior module is detached during testing, ZID-Net achieves processing speeds of 5.14 ms and 19.35 ms, ranking as the fastest among all evaluated methods. This efficiency becomes particularly evident when compared to the standard diffusion-based method, *Learning Hazing to Dehazing* [29], which demands 9540.21 ms and 19870.79 ms per image due to iterative reverse sampling. Ultimately, our zero-inference deterministic design overcomes the inherent latency bottlenecks of diffusion models, enabling real-time inference without compromising restoration fidelity.

**Table 4**

Comparison of computational efficiency among representative dehazing models evaluated on 256 × 256 images.

| Method | FFA | SGID | C2PNet | Dehaze Former-B | SANet | DCMPNet | DEA-Net | MB-Taylor FormerV2-B | TUR | SGDN | *Learning Hazing to Dehazing* | Ours |
|---|---|---|---|---|---|---|---|---|---|---|---|---|
| Params (M) | 4.46 | 13.87 | 7.17 | 2.52 | 3.81 | 17.36 | 3.65 | 2.63 | 2.60 | 11.08 | 1228.89 | 17.51 |
| FLOPs (GMac) | 287.53 | 156.40 | 460.95 | 29.63 | 44.78 | 68.83 | 34.04 | 31.367 | 17.03 | 52.95 | 114.35 | 17.91 |
| Runtime (ms) | 78.02 | 49.86 | 112.71 | 54.18 | 48.82 | 34.35 | 11.35 | 132.03 | 22.62 | 53.85 | 9540.21 | 5.14 |

**Table 5**

Comparison of computational efficiency among representative dehazing models evaluated on 512 × 512 images.

| Method | FFA | SGID | C2PNet | Dehaze Former-B | SANet | DCMPNet | DEA-Net | MB-Taylor FormerV2-B | TUR | SGDN | *Learning Hazing to Dehazing* | Ours |
|---|---|---|---|---|---|---|---|---|---|---|---|---|
| Params (M) | 4.46 | 13.87 | 7.17 | 2.52 | 3.81 | 17.36 | 3.65 | 2.63 | 2.60 | 11.08 | 1228.89 | 17.51 |
| FLOPs (GMac) | 1150.13 | 625.61 | 1843.82 | 93.99 | 188.08 | 275.08 | 136.17 | 125.47 | 68.01 | 211.86 | 448.63 | 71.64 |
| Runtime (ms) | 479.2 | 189.12 | 682.64 | 253.26 | 219.25 | 155.87 | 51.94 | 536.15 | 101.38 | 224.63 | 19870.79 | 19.35 |

## 4.6. Ablation Study

To evaluate the design choices of ZID-Net, we conduct ablation studies on the Real-World, StateHaze1k, and SOTS-Outdoor datasets. We first analyze the contributions of the main network components and compare ZI-PPH with other auxiliary supervision heads. We then examine the correction fields produced by these heads, compare different timestep sampling strategies, evaluate different high-frequency extraction operators, and study the effects of the training losses and their weights.

**Effectiveness of Network Components.** As shown in Table 6, the full ZID-Net achieves the best performance across all testing scenarios, with the corresponding visual results for the Real-World Thin dataset presented in Fig. 11. Removing the structural branch leads to substantial PSNR decreases of 2.61 dB on the Real-World Thin dataset and 2.18 dB on the Real-World Dense dataset. As illustrated in Fig. 11, this quantitative decline is accompanied by noticeable structural blurring, emphasizing the critical role of explicitly modeling high-frequency physical residuals. Similarly, discarding LGCBs or replacing DFAB with simple concatenation degrades metrics and reduces visual contrast across both datasets. Notably, training without the ZI-PPH branch reduces the PSNR from 21.83 dB to 19.89 dB under thin haze and from 20.87 dB to 19.27 dB under dense haze. These metric losses, alongside the residual haze visible in the ablation images in Fig. 11, demonstrate that the zero-inference diffusion prior provides essential structural regularization for the shared bottleneck.

**Table 6**

Ablation study on the architectural components of ZID-Net. The evaluation is conducted on the Real-World Thin and Real-World Dense datasets. Best results are highlighted in bold.

| Method | Real-World Thin | | Real-World Dense | |
|---|---|---|---|---|
| | PSNR↑ | SSIM↑ | PSNR↑ | SSIM↑ |
| w/o Structural Branch | 19.22 | 0.658 | 18.69 | 0.495 |
| w/o ZI-PPH | 19.89 | 0.701 | 19.27 | 0.534 |
| w/o LGCBs | 20.91 | 0.727 | 19.86 | 0.576 |
| w/o DFAB | 20.10 | 0.741 | 19.52 | 0.579 |
| Full | **21.83** | **0.776** | **20.87** | **0.601** |

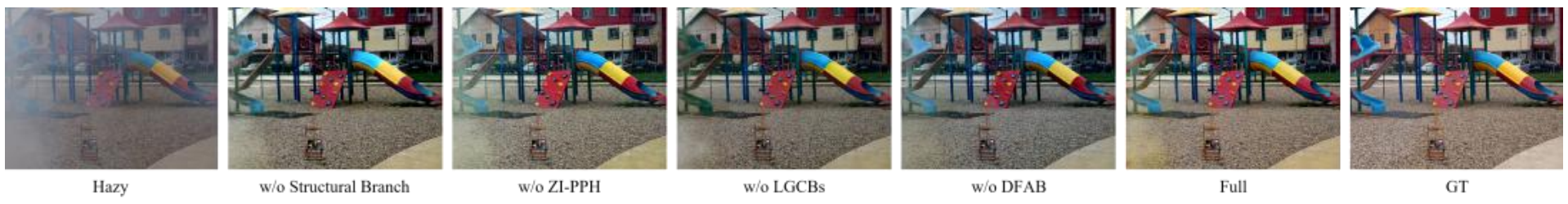

**Fig. 11.** Visual ablation results of ZID-Net on the Real-World Thin dataset.

**Comparison of Auxiliary Supervision.** We attach four alternative prediction heads to the shared bottleneck to compare different forms of auxiliary supervision. Following the atmospheric scattering model, the $A$-Head predicts global atmospheric light ($A$) as a three-dimensional color vector, while the $t$-Head estimates a single-channel transmission map ($t$). The joint $A + t$ Head combines both predictions. The Residual-Head instead directly reconstructs the three-channel haze residual. Table 7 compares the restoration performance of the resulting variants, among which ZI-PPH achieves the highest quantitative results. To further investigate the reason for this performance difference, we conduct a correction-field analysis. As shown in Fig. 12, for each trained variant $m$, we derive its correction field as $\Delta I_m = \hat{I}_m - I_h$, where $\hat{I}_m$ denotes the restored output of the corresponding variant. The reference correction field is defined as $\Delta I_{\mathrm{GT}} = I_c - I_h$. Consistent with the results in Table 7, ZI-PPH produces a correction field that is closest to the reference, achieving the highest spatial correlation and the lowest MAE. In comparison, the simpler auxiliary heads produce less spatially consistent correction patterns. These results suggest that diffusion-based noise-prediction supervision provides a more effective training signal for spatially varying haze than direct physical-parameter or residual regression. Since all auxiliary heads are removed during testing, these improvements introduce no additional inference cost.

**Table 7**

Quantitative comparison of different auxiliary supervision heads during training on the Real-World Thin dataset. All auxiliary heads are detached during inference. Best results are highlighted in bold.

| **Method** | **Real-World Thin** | | |
|---|---|---|---|
| | **PSNR↑** | **SSIM↑** | **Training overhead (s/batch) ↓** |
| Backbone only | 19.89 | 0.701 | **1.02** |
| + $A$-Head | 19.92 | 0.709 | 1.10 |
| + $t$-Head | 19.71 | 0.711 | 1.15 |
| + $A + t$-Head | 20.68 | 0.724 | 1.16 |
| + Residual Head | 20.43 | 0.708 | 1.16 |
| + ZI-PPH (Ours) | **21.83** | **0.776** | 1.21 |

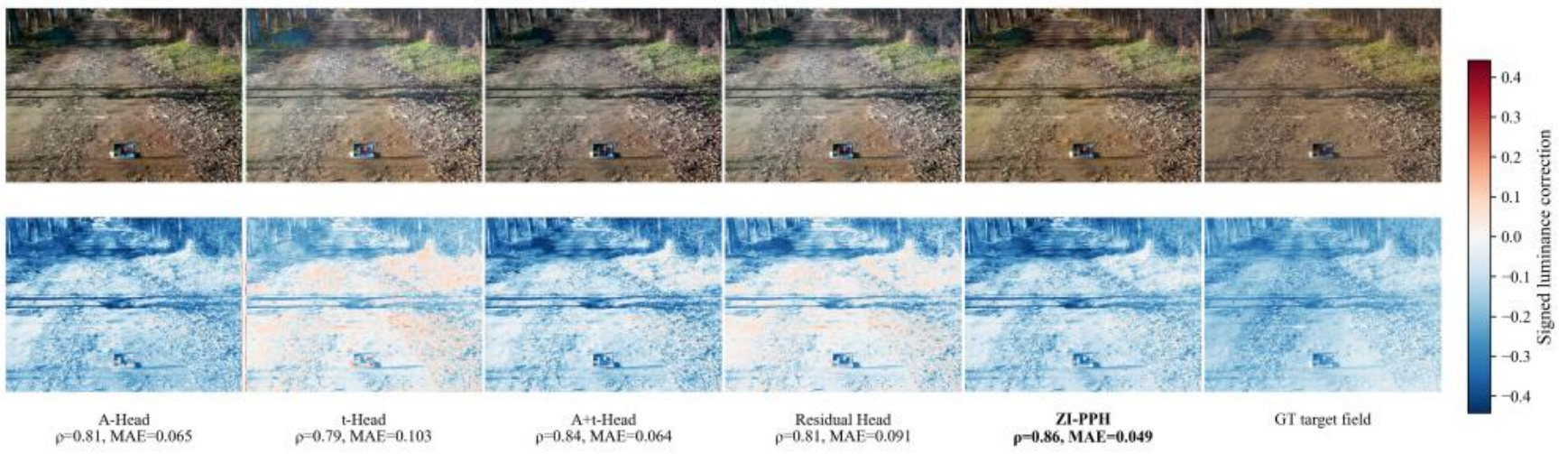

**Fig. 12.** Correction-field comparison of different auxiliary-supervision variants. $\rho$ denotes the spatial correlation coefficient, and MAE denotes the mean absolute error; both are computed against the reference correction field $\Delta I_{\mathrm{GT}}$. Higher $\rho$ and lower MAE indicate better agreement.

**Effect of Timestep Sampling.** We further compare the proposed adaptive strategy with uniform timestep sampling on StateHaze1k. As shown in Table 8, adaptive sampling improves all three haze levels, increasing the average PSNR by 1.25 dB and SSIM by 0.013. Unlike uniform sampling, it assigns smaller timestep ranges to light-haze samples to avoid excessive corruption of fine details, while exposing heavier-haze samples to a broader noise range to strengthen structural modeling. This better matches the perturbation strength to the actual degradation level and provides more effective supervision for the shared bottleneck.

**Table 8**

Effect of timestep sampling on StateHaze1k. The best result in each column is highlighted in bold.

| Sampling strategy | Thin | | Moderate | | Thick | |
|---|---|---|---|---|---|---|
| | PSNR↑ | SSIM↑ | PSNR↑ | SSIM↑ | PSNR↑ | SSIM↑ |
| Uniform timestep sampling | 24.40 | 0.920 | 25.95 | 0.936 | 21.76 | 0.849 |
| Adaptive timestep sampling (Ours) | **25.26** | **0.928** | **27.71** | **0.947** | **22.91** | **0.869** |

**Analysis of Structural Prior Inputs.** As detailed in Table 9, the proposed Color Laplacian achieves the highest PSNR and SSIM alongside the lowest $\Delta E_{ab}$ [42] and $\Delta E_{00}$ [43] color distortion metrics. This distinct advantage stems from preserving signed cross-channel variations, which alternative extractors inherently destroy. Canny and the Gray Laplacian collapse the input into a single luminance channel, eliminating crucial cross-channel degradation differences. Operators such as Sobel, Prewitt, and Scharr process independent RGB channels. However, they compute non-negative gradient magnitudes and inherently discard critical directional signs. This physical limitation is also shared by Gabor filters, which compress high-frequency representations via absolute averaging. The Color Laplacian circumvents these issues by computing signed second-order residuals directly across the RGB channels, explicitly capturing both directional transition signs and color-specific attenuation priors. The practical impact of retaining these uncompressed structural and chromatic details is visually evident in Fig. 13.

**Table 9**

Quantitative comparison of different high-frequency operators for the structural prior. Best results are highlighted in bold.

| Method | Real-World Thin | | | | SOTS-Outdoor | | | |
|---|---|---|---|---|---|---|---|---|
| | PSNR↑ | SSIM↑ | $\Delta E_{ab}$ ↓ | $\Delta E_{00}$ ↓ | PSNR↑ | SSIM↑ | $\Delta E_{ab}$ ↓ | $\Delta E_{00}$ ↓ |
| Canny | 18.17 | 0.634 | 14.025 | 11.199 | 34.53 | 0.993 | 2.093 | 1.557 |
| Gabor | 17.79 | 0.674 | 15.803 | 12.448 | 35.08 | 0.980 | 2.368 | 1.900 |
| Sobel | 17.59 | 0.613 | 14.562 | 11.666 | 35.88 | 0.986 | 1.847 | 1.406 |
| Prewitt | 17.58 | 0.613 | 14.571 | 11.674 | 35.88 | 0.986 | 1.848 | 1.407 |
| Scharr | 17.61 | 0.615 | 14.569 | 11.672 | 35.83 | 0.986 | 1.847 | 1.403 |
| Gray Laplacian | 21.74 | 0.761 | 10.446 | 8.476 | 38.25 | 0.992 | 1.516 | 1.143 |
| Color Laplacian (Ours) | **21.83** | **0.776** | **10.401** | **8.409** | **38.73** | **0.993** | **1.408** | **1.078** |

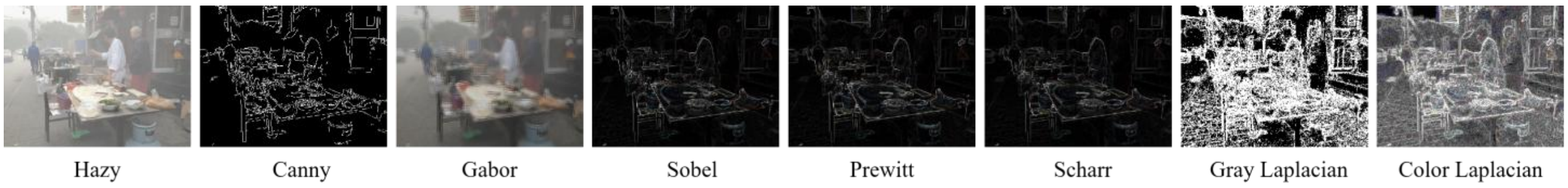

**Fig. 13.** Visual comparison of high-frequency features extracted by different operators.

**Ablation of Loss Functions and Weights.** Table 10 evaluates the contributions and balance of the two auxiliary losses, with the reconstruction weight kept at 1.0. Removing the contrastive loss decreases PSNR by 1.12 dB and SSIM by 0.056, while removing the diffusion loss causes larger drops of 1.71 dB and 0.092, respectively. This indicates that diffusion supervision provides the stronger constraint for learning haze-related residual distributions, whereas contrastive supervision offers complementary feature discrimination. The weight analysis shows a non-monotonic effect: overly weak contrastive supervision cannot provide sufficient feature guidance, while an excessive weight competes with image reconstruction. Similarly, insufficient or excessive diffusion supervision disturbs the balance between residual modeling and restoration. Among the tested settings, the adopted weights provide the most balanced restoration performance.

**Table 10**

Ablation of loss terms and weights on the Real-World Dense dataset. $\lambda_1$ is fixed to 1.0. Best available values are highlighted in bold.

| $\lambda_2$ | $\lambda_3$ | Real-World Dense | | |
|---|---|---|---|---|
| | | **PSNR↑** | **SSIM↑** | **$\Delta E_{00}$↓** |
| 0.00 | 0.35 | 19.75 | 0.545 | 13.32 |
| 0.10 | 0.00 | 19.16 | 0.509 | 13.36 |
| 0.05 | 0.35 | 19.70 | 0.558 | 13.09 |
| 0.20 | 0.35 | 18.70 | 0.531 | 14.07 |
| 0.10 | 0.20 | 18.80 | 0.510 | 14.48 |
| 0.10 | 0.50 | 18.65 | 0.510 | 14.38 |
| 0.10 | 0.35 | **20.87** | **0.601** | **10.83** |

## 5. Conclusion

In this paper, we proposed ZID-Net, a novel frequency-spatial decoupled framework that addresses the conflict between the generative capability of diffusion models and the latency requirements of low-level vision tasks. During training, the Zero-Inference Prior Propagation Head introduces dynamic noise prediction across varying timesteps, providing degradation-aware supervision for the shared bottleneck. This diffusion branch is completely removed during inference, allowing the network to benefit from diffusion-based regularization without introducing additional computational overhead. To establish long-range spatial dependencies, Lightweight Global Context Blocks are incorporated into the backbone, while a parallel high-frequency branch employs the Channel-Spatial Laplacian Mask to suppress haze-amplified responses and preserve chromatic and structural details. These complementary representations are integrated through the Dynamic Feature Arbitration Block, which adaptively reweights semantic, structural, and upsampled features during image reconstruction. Extensive experiments on synthetic, real-world, and remote-sensing datasets demonstrate that ZID-Net achieves competitive restoration performance while maintaining deterministic and efficient inference. The current design does not explicitly adapt to unseen data distributions during inference, and its performance under substantial domain shifts, nighttime illumination changes, and near-complete scene occlusion remains to be further investigated. Future work will explore more general degradation representations and cross-domain restoration strategies to further improve robustness under diverse adverse conditions.